\documentclass[journal]{IEEEtran}
\usepackage[T1]{fontenc}
\usepackage{mathptmx}

\usepackage{times}
\usepackage{epsfig}
\usepackage{graphicx}
\usepackage{amsmath}
\usepackage{amssymb}
\usepackage[pagebackref=true,breaklinks=true,letterpaper=true, colorlinks, bookmarks=false]{hyperref}

\begin{document}
\title{\fontfamily{ptm}\selectfont \fontsize{18}{20}\selectfont Fine-grained Visual Classification with High-temperature Refinement and Background Suppression}

\author{
\IEEEauthorblockN{Po-Yung Chou,
                  Yu-Yung Kao, and
                  Cheng-Hung Lin*\thanks{ *corresponding author}}\\
\IEEEauthorblockA{Department of Electrical Engineering, National Taiwan Normal University}\\
\IEEEauthorblockA{\tt\small \{81075001H, 81075006h, brucelin\}@ntnu.edu.tw}
}

\markboth{March~2023}%
{Shell \MakeLowercase{\textit{et al.}}: Bare Demo of IEEEtran.cls for IEEE Journals}

\maketitle
\begin{abstract}
Fine-grained visual classification is a challenging task due to the high similarity between categories and distinct differences among data within one single category. To address the challenges, previous strategies have focused on localizing subtle discrepancies between categories and enhencing the discriminative features in them. However, the background also provides important information that can tell the model which features are unnecessary or even harmful for classification, and models that rely too heavily on subtle features may overlook global features and contextual information. In this paper, we propose a novel network called ``High-temperaturE Refinement and Background Suppression'' (HERBS), which consists of two modules, namely, the high-temperature refinement module and the background suppression module, for extracting discriminative features and suppressing background noise, respectively. The high-temperature refinement module allows the model to learn the appropriate feature scales by refining the features map at different scales and improving the learning of diverse features. And, the background suppression module first splits the features map into foreground and background using classification confidence scores and suppresses feature values in low-confidence areas while enhancing discriminative features. The experimental results show that the proposed HERBS effectively fuses features of varying scales, suppresses background noise, discriminative features at appropriate scales for fine-grained visual classification.The proposed method achieves state-of-the-art performance on the CUB-200-2011 and NABirds benchmarks, surpassing 93\% accuracy on both datasets. Thus, HERBS presents a promising solution for improving the performance of fine-grained visual classification tasks. code will be available: \url{https://github.com/chou141253/FGVC-HERBS}
\end{abstract}
\section{Introduction}

\IEEEPARstart
{F}{ine-grained} visual classification (FGVC) is a challenging task in computer vision that involves categorizing images into very specific and detailed categories, such as different species of birds\cite{CUB_200_2011}, dogs\cite{Stanford_Dogs}, vehicle models\cite{Stanford_Cars}, and medical images\cite{FGRDA}. As shown in Fig.\ref{fig:ex_coarse_fine}, these four types of sparrows look almost identical, but from different perspectives, the same type of sparrow also looks very different. 
In contrast to coarse-grained classification, which involves identifying broad categories like ``animals" or ``vehicles," fine-grained classification requires the ability to recognize subtle differences in visual features, such as color, texture, shape, and pattern, which often exist in small regions. These regions are referred to as discriminative regions or foreground regions.

Fine-grained recognition can be achieved by dividing objects into parts, such as eyes, feet, etc., and comparing corresponding regions for easier identification\cite{HPM,HOG_Alignments,TruesAboutDog, PB-RCNN,PS-CNN,PNDCN,SPDA-CNN}. However, these methods require manual annotation, which is costly and even requires expert annotation. To overcome this issue, the weakly supervised methods\cite{Granularity-Specific, Re-rank-Local, SAC, WS-DAN, S3Ns, MMP-ScPM, LR2M} are proposed to find discriminative regions through class activation mapping (CAM)\cite{CAM, Grad-CAM} were proposed, offering training the network through higher response areas in the feature map without labels.
 In addition, the attention-based methods\cite{MA-CNN, DCAL, API-Net, PCA-Net, BNT} are proposed to locate discriminative regions by identifying common high-response areas among feature maps. Furthermore, the success of Vision Transformer (ViT) in image classification has led to its implementation in fine-grained visual recognition tasks. These methods\cite{SACM, SIM-TRans, TransFG, FFVT, Vit-KG-PS, RAMS-Trans} use self-attention maps to get information on foreground regions. The main efforts were focused on enhancing the differentiation of discriminative regions, while neglecting unselected regions. However, in cases where the model cannot obtain strong enough discriminative regions, it is useful to first exclude unimportant regions, called background. Motivated by this concept, we propose the Background Suppression (BS) module.

The proposed BS module shows better performance in FGVC tasks. In the first step of the BS module, the output confidence scores are utilized to classify regions into foreground and background. The foreground represents the discriminative area, while the background refers to the unselected or noisy part. Subsequently, the BS module suppresses the feature values in low confidence regions and enhances the discriminative features, thus improving the details of the target object and reducing the noise. Therefore, the BS module can be helpful, especially in cases where it is difficult to distinguish between foreground and background areas.

The algorithm for extracting features from discriminative regions is important for the FGVG task. However, it can lead to the problem of losing contextual information due to the overuse of single or few specific categories of features. Therefore, we propose a high-temperature refinement module to enhance the learning of diverse features, including texture, shape, and appearance from various object categories. Specifically, the module initially uses higher temperatures to learn feature maps so that more global and contextual information can be captured. Subsequently, the feature maps were refined using lower temperatures to capture finer details. This approach allows to obtain richer features, to better classify similar objects, and to improve accuracy. It should be noted that the high-temperature refinement module can be considered as a form of knowledge distillation \cite{KD}.

The high-temperature refinement module also maintains an appropriate size of the discriminative region, which is advantageous for FGVC tasks. If the feature size is too small, the algorithm may not be able to capture the overall features of the object, resulting in incorrect classification. Conversely, if the feature scale is too large, the accuracy of FGVC tasks may be reduced due to excessive noise and redundant information.

In this paper, the proposed \textbf{H}igh temperatur\textbf{E} \textbf{R}efinement and \textbf{B}ackground \textbf{S}uppression (\textbf{HERBS}) can extract discriminative features and suppress background noise. This paper has two main contributions:
\begin{itemize}
\item  The proposed HERBS can be integrated into various backbones, such as CNN-based networks and transformer-based networks. It also allows to perform end-to-end training.
\item The proposed HERBS outperforms state-of-the-art approaches, improving the accuracy to 93.1\% and 93.0\% on CUB200-2011\cite{CUB_200_2011} and NABirds\cite{NA-Birds}, respectively.
\begin{figure}[t]
    \begin{center}
    \includegraphics[width=1.0\linewidth]{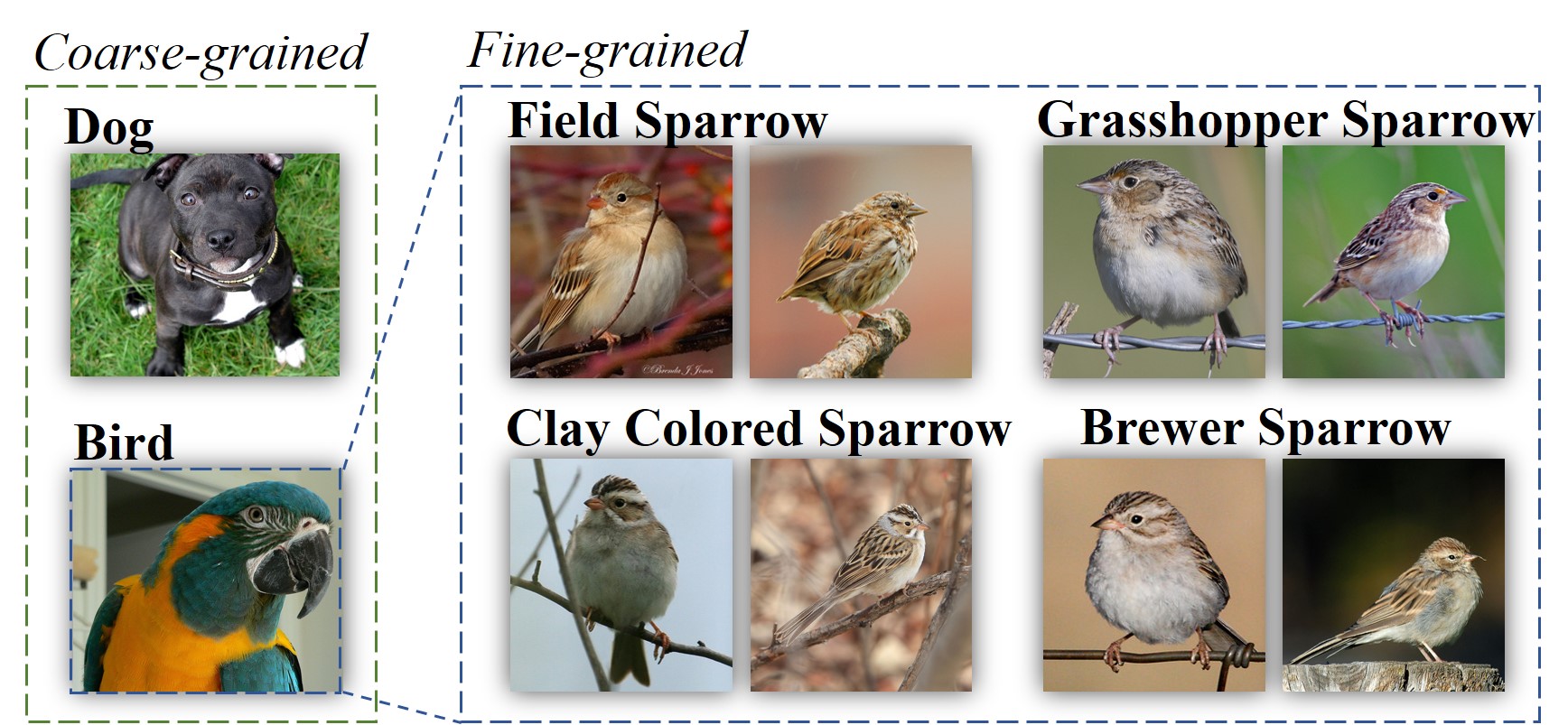}
    \end{center}
\caption{Examples of visual classification for coarse-grained categories and fine-grained categories.}
\label{fig:ex_coarse_fine}
\end{figure}
\end{itemize}
\section{Related Work}

\subsection{Fine-grained visual classification}
In the field of FGVC, there are two approaches for extracting discriminative features from subtle areas, broadly classified as object-part-based methods and attention-based methods.

\textbf{Object-part based methods} aim to find object local areas for recognition by using a model to generate candidate regions, then extracting discriminative features from them. MA-CNN\cite{MA-CNN} trains positioning and classification accuracy at the same time through clustering of feature maps into object parts. This unsupervised classification enhances feature learning by dividing patterns into object parts. The approach allows for simultaneous learning of discriminative features and positions. S3N\cite{S3Ns} finds the local extremes of each category response on the feature map to enhance features. In addition, WS-DAN\cite{WS-DAN} augment the data by cutting out local extremes to discover other discriminative features.

\textbf{Attention-based methods}, on the other hand, use attention mechanisms to enhance feature learning and locate object details. MAMC\cite{MAMC} generates multiple sets of features enhanced by attention mechanisms, Cross-X\cite{Cross-X} 
use attention maps from multi-excitation models to learn features from different caterories. API-Net\cite{API-Net} and PCA-Net\cite{PCA-Net} uses two images as input to calculate attention between feature maps to enhence discriminative representations. CAP\cite{CAP} calculates the self-attention map of the output features to express the relationship between feature pixels, and SR-GNN\cite{SR-GNN} uses graph convolutional neural networks to describe the relationship between parts. CAL\cite{CAL} adds a counterfactual intervention to the attention map to predict the category. With the development of Transformer\cite{Transformer} in the computer vision field, many improved Vision Transformer architectures have been proposed, such as FFVT\cite{FFVT}, SIM-Trans\cite{SIM-TRans}, TransFG\cite{TransFG}, and AFTrans\cite{AFTrans}, these methods utilize self-attention maps in transformer layers to enhance feature learning and locate object details.

\subsection{Object detection}
Supervised object detection methods have demonstrated significant results. The supervised YOLOv7\cite{YOLOv7} can achieve fast and high accuracy of detection. However, the manual labeling requirement for object positions limits its suitability for fine-grained visual recognition tasks.

Weakly supervised object detection (WSOD) has been introduced as an alternative to overcome the limitations. This method only requires classification labels and generates pseudo bounding box targets through algorithms. For instance, WCCN\cite{WCCN} generates class activation maps to identify regions of interest, which are then fed into the classifier and corrected through multiple instance learning. WSOD2\cite{WSOD2} scores virtual candidate boxes through Top-Down and Bottom-Up approaches, with the virtual box with the highest score serving as the target output for the next layer. MIST\cite{MIST} refines regions of interest through self-training, while WSCL\cite{WSCL} improves the features of regions of interest through data enhancement and contrastive learning. These methods gradually discover the whole object through refinement processes, utilizing the output of the previous stage as the virtual target.
\section{Method}

\begin{figure}[t]
    \begin{center}
    \includegraphics[width=1.0\linewidth]{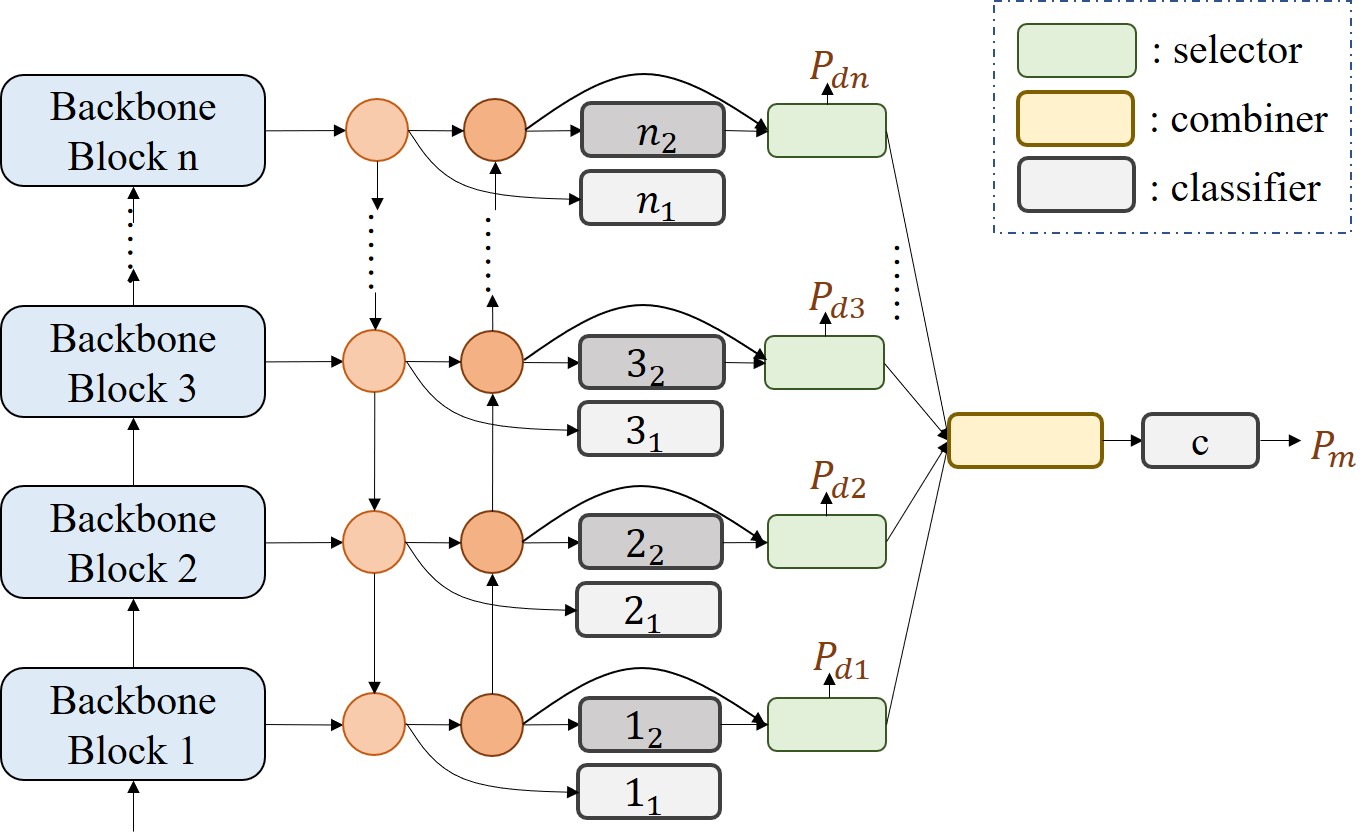}
    \end{center}
\caption{The illustration of the model structure is shown, where the blue squares on the left represent the backbone blocks, which could be either Convolution-based or Transformer-based. The circles in the middle part denote the multi-scale feature fusion module, such as Feature Pyramid Network (FPN) or Path Aggregation (PA). The classifier, selector, and combiner on the right side depict the HERBS module.}
\label{fig:HERBS_structure}
\end{figure}

In Fig.\ref{fig:HERBS_structure}, the proposed High-temperatureE Refinement and Background Suppression (HERBS) network is composed of the backbones, the top-down features fusion module, the bottom-up features fusion module, and the HERBS. The backbone can be either a Transformer-based model (e.g., Swin Transformer) or a Convolution-based model (e.g., ResNet). The top-down and bottom-up features fusion module is similar to the path aggregation network (PA)\cite{PANet}, which can be treated as a feature pyramid network (FPN)\cite{FPN} with an additional bottom-up path. 

The proposed HERBS networks aims to learn diverse and discriminative features and improve the accuracy of several FGVC tasks. HERBS contains two modules: the background suppression (BS) module and the high-temperature refinement module. In the following sections, we refer to these two fusion modules as the top-down path and the bottom-up path. And the proposed HERBS, we will provide a comprehensive description of the design of the BS module and high-temperature module, including a detailed explanation of the use of the loss function and the integration of the HERBS module with various frameworks.

\subsection{Background suppression}

Let $hs_{i}$ denote the features map generated by the $i^{th}$ backbone block, where $hs_{i} \in R^{C_{i} \times H_{i} \times W_{i}}$. Here, $C_{i}$ represents the number of channels, $H_{i}$ is the height, and $W_{i}$ is the width of the features map. The first step of the background suppression (BS) module is to generate the classification maps from these features map, which can be expressed as:
\begin{equation} \label{eq:11}
Y_{i} = W_{i}hs_{i} + b_{i}
\end{equation}
where $W_{i}$ is the weight of the $i^{th}$ layer classifier, $b_{i}$ is its bias, and $Y_{i}$ is the classification maps, with dimensions $R^{C_{gt}\times H_{i}\times W_{i}}$, where $C_{gt}$ is the number of target categories. Then the maximum score map is calculated from classification map. The process can be expressed as:
\begin{equation} \label{eq:12}
P_{max, i} = \text{max}(\text{Softmax}(Y_{i}))
\end{equation}
where $P_{max, i}$ represents the i-$th$ layer's max score map. Next, the features with the top-$K_{i}$ scores among all predictions are selected. The number of $K_{i}$ is selected based on the principle that $K_{i} > K_{j}$ when $i < j$. Specifically, we set $K_{1}$ to 256, $K_{2}$ to 128, $K_{3}$ to 64, and $K_{4}$ to 32. We select this value based on the principle that earlier layers can limit the performance of subsequent layers, and our experiments show that the accuracy is relatively insensitive to variations in this parameter if this principle is followed.

A graph convolution module is then employed to merge the selected features and make a prediction based on the merged features. At this stage, the BS module has the non-selected classification maps, referred to as the dropped maps, denoted as $Y_{d}$, and the merged classification prediction, denoted as $Y_{m}$. This process is depicted through the selector and combiner components as shown in Fig.\ref{fig:HERBS_structure}. 

The objective function of the merged classification prediction is a standard classification one, using cross-entropy to calculate the similarity between the prediction distribution $P_{m}$ and the ground truth label $y$. The merged loss is calculated as follows:
\begin{equation} \label{eq:p_m}
P_{m} = \text{Softmax}(Y_{m})
\end{equation}
\begin{equation} \label{eq:cross_entropy}
loss_{m} = -\sum_{ci=1}^{C_{gt}} y_{ci} \log(P_{m,ci})
\end{equation}

Here, $y_{ci}$ is the ground truth of $i^{th}$ class, and $P_{m,ci}$ is the predicted probability of the $i^{th}$ class. The summation is performed over the number of target categories $C_{gt}$. This enhances the discriminative features in the selected area.  

The other objective of the BS module is to suppress features in the dropped maps and increase the gap between the foreground and background. The hyperbolic tangent function, $\text{tanh}$, is applied to the dropped maps, $Y_{d}$, as shown in Eq.\eqref{eq:p_d}:
\begin{equation} \label{eq:p_d}
P_{d} = \text{tanh}(Y_{d})
\end{equation}

Then, the dropped loss, $loss_{d}$, is calculated as the mean squared error between the prediction and a pseudo target of -1, as defined in Eq.\eqref{eq:mse_loss}:
\begin{equation} \label{eq:mse_loss}
loss_{d} = \sum_{i=1}^{C_{gt}} (P_{d,ci} + 1)^2
\end{equation}

Note that the hyperbolic tangent function in Eq. \eqref{eq:p_d} maps the values of the prediction into a range that is not restricted to probabilities. This is because we really want to separate foreground and background features even if the background have some other classes appearances. 

In order to prevent all blocks' feature maps from only having high responses in the same locations, we also incorporate the prediction of each layer into the training target as follows:
\begin{equation} \label{eq:layer_pred}
P_{li} = \text{Softmax}(W_{i}(\text{Avgpool}(hs_{i})) + b_{i})
\end{equation}
\begin{equation} \label{eq:loss_l}
loss_{l} = -\sum_{i=1}^{n} \sum_{ci=1}^{C_{gt}} y_{ci} \log(P_{li, ci})
\end{equation}
where $\text{Avgpool}$ function aggregates all $H_{i}$, and $W_{i}$ at each channel, and the number of blocks in the backbone is represented by $n$.

The total BS objective is given by the weighted sum of the merged loss ($loss_{m}$), dropped loss ($loss_{d}$), and average layer loss ($loss_{l}$), as shown in Eq.\eqref{eq:bs_loss}:
\begin{equation} \label{eq:bs_loss}
loss_{bs} = \lambda_{m}loss_{m} + \lambda_{d}loss_{d} + \lambda_{l}loss_{l}
\end{equation}
where $\lambda_{m}$, $\lambda_{d}$, and $\lambda_{l}$ are the weights for the merged loss, dropped loss, and average layer loss, respectively. Specifically, We set $\lambda_{m}$ to 1, $\lambda_{d}$ to 5, and $\lambda_{l}$ to 0.3. These values were set to balance the foreground and background loss and were determined based on the training loss from the first three epochs.

\subsection{High-temperature refinement}

\begin{figure}[t]
    \begin{center}
    \includegraphics[width=1.0\linewidth]{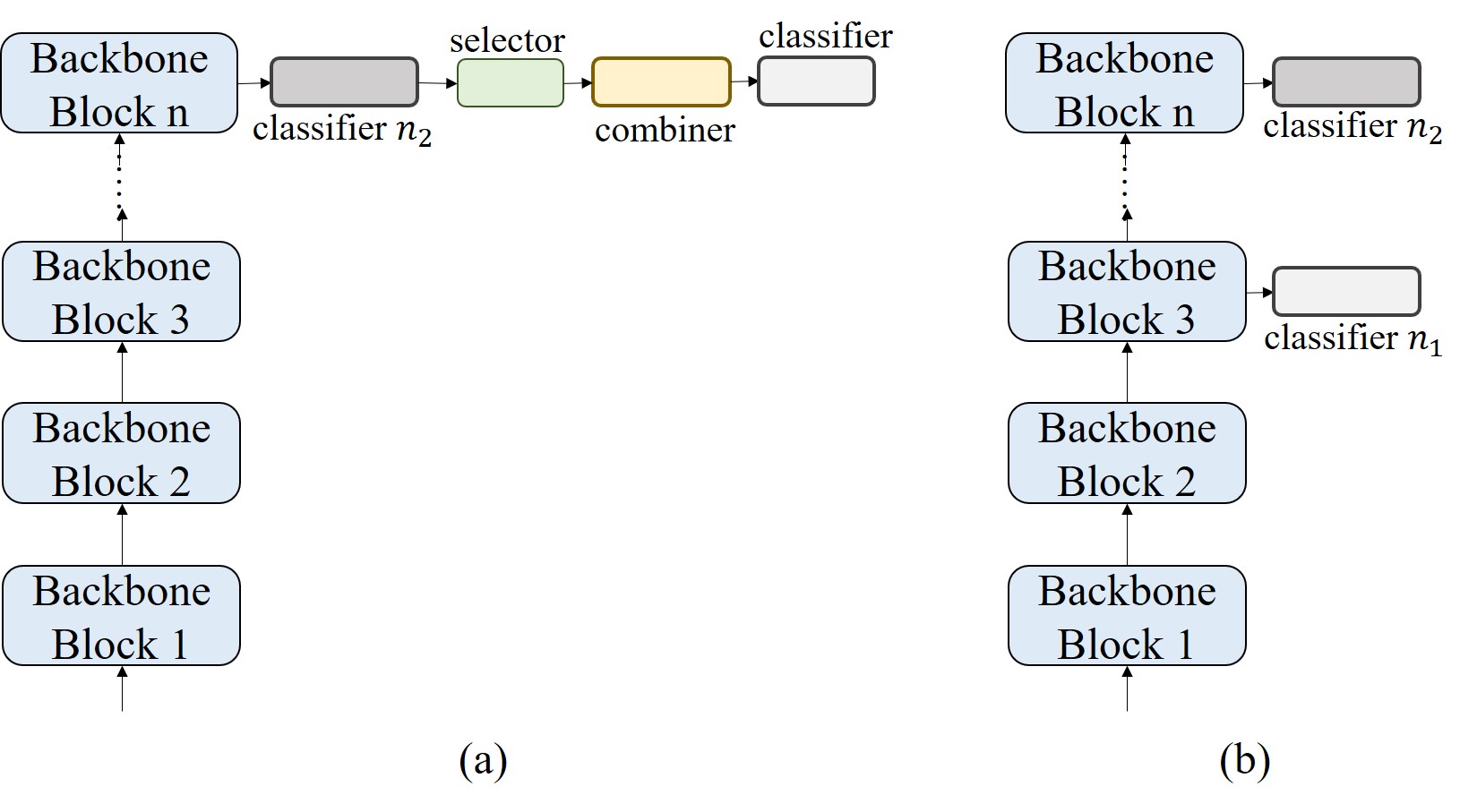}
    \end{center} 
\caption{Illustration of the structure of (a) basic background suppression module and (b) basic high-temperature refinement module.}
\label{fig:basic_module}
\end{figure}

The classifier $k_{1}$ and classifier $k_{2}$ in the Fig.\ref{fig:HERBS_structure} are followed by the $k$-th block features map, classifier $k_{1}$ is located within the top-down path while classifier $k_{2}$ is in the bottom-up one. The objective is to make classifier $k_{1}$ learn the output distribution of classifier $k_{2}$. We define the output of classifier $k_{1}$ as $Y_{i1}$ and the output of classifier $k_{2}$ as $Y_{i2}$. The refinement objective function helps the model to learn more diverse and stronger representations in the earlier layers while allowing later layers to focus on finer details. In other words, the high-temperature refinement module enables classifier $k_{1}$ to discover broader areas and classifier $k_{2}$ to focus on learning fine-grained and discriminative features. The refinement loss is calculated using the following equations:
\begin{equation} \label{eq:P_i1}
P_{i1} =  \text{LogSoftmax}(Y_{i1} / T_{e})
\end{equation}
\begin{equation} \label{eq:P_i2}
P_{i2} =  \text{Softmax}(Y_{i2} / T_{e})
\end{equation}
\begin{equation} \label{eq:kl_loss}
loss_{r} =  P_{i2}\log(\frac{P_{i2}}{P_{i1}})
\end{equation}
where $T_{e}$ represents the temperature at training epoch $e$. The value of $T_{e}$ decreases as the training epoch increases, following a decay function defined as:
\begin{equation} \label{eq:decrease_func}
T_{e} = 0.5^{^{\left\lfloor\frac{e}{-\log_{2}(0.0625/T)}\right\rfloor}}
\end{equation}

We set the initial temperature $T$ to a high value, such as 64 or 128, in comparison to the knowledge distillation approach\cite{KD}. The aim is to encourage the model to explore various features even if the initial predictions are inaccurate. Then, as training progresses, the temperature gradually decreases, allowing the model to focus more on the target class and learn more discriminative features. By using this decay policy, the model can obtain diverse and fine representations and make accurate predictions.

The total loss of HERBS can be formulated as:
\begin{equation} \label{eq:herbs_loss}
loss_{herbs} = loss_{bs} + \lambda_{r}loss_{r}
\end{equation}
where $\lambda_{r}$ is the weight for refinement loss, which set to 1. And the HERBS network's final output is the softmax of the sum of nine classifier results, consisting of four from the top-down approach, four from the bottom-up approach, and one from the combiner.

Note that in the HERBS network, $W_{i}$ and $b_{i}$ belong to classifier $k_{2}$ when $i$ equals to $k$. We separately describe them because the BS module and high-temperature refinement module can be applied to the backbone alone, which is very flexible. The experimental results show that both modules can improve accuracy. Of course, when using the entire HERBS network, the model's capability will result in even better performance.

In this paper, we propose the HERBS module, composed of the first Background Suppression (BS) and the high-temperature refinement module. Both components can improve the backbone model's accuracy in FGVC tasks. We show the most basic BS and high-temperature refinement modules in Fig.\ref{fig:basic_module}(a) and (b), respectively. The most basic BS module is added to the output of the final block and to achieve Eqs.\eqref{eq:11}--\eqref{eq:bs_loss}. And the most basic High-temperature refinement module is applied to the last two blocks. The final classifier would be treated as classifier $n_{2}$ and another one as classifier $n_{1}$. Following Eq\ref{eq:kl_loss}, we calculate their KL-divergence as the objective function.

\section{Experiments}

\begin{table}
\begin{center}
\begin{tabular}{|l|c|c|}
\hline
Method & CUB-200-2011 & NA-Birds\\
\hline
FFVT\cite{FFVT} & 91.6 & N/A\\
ViT-NeT\cite{Vit-NeT} & 91.7 & N/A \\
TransFG\cite{TransFG} & 91.7 & 90.8 \\
IELT\cite{IELT} & 91.8 & 90.8 \\
SIM-Trans\cite{SIM-TRans} & 91.8 & N/A\\
SAC\cite{SAC} & 91.8 & N/A\\
CAP\cite{CAP} & 91.9 & 91.0\\
SR-GNN\cite{SR-GNN} & 91,9 & 91.2\\
DCAL\cite{DCAL} & 92.0 & N/A \\
MetaFormer\cite{MetaFormer} & 92.4 & 92.7\\
\hline
HERBS & \textbf{93.1}  & \textbf{93.0}\\
\hline
\end{tabular}
\end{center}
\caption{Comparison of top-1 accuracy(\%) with state-of-the-art methods on the two benchmarks, CUB-200-2011 and NA-Birds.}
\label{tab1}
\end{table}

\begin{table}
\begin{center}
\begin{tabular}{|c|c|c|c|c|}
\hline
\multicolumn{3}{|c|}{Module} & \multicolumn{2}{|c|}{Backbone} \\
\hline
PA & Refinement & BS & Swin-Base & Swin-Large\\
\hline
&  &  & 91.3 & 92.0\\
\hline
\checkmark &  &  & 91.9(+0.6) & 92.5(+0.5)\\
& \checkmark &  & 91.5(+0.2) & 92.3(+0.3)\\
&  & \checkmark & 91.8(+0.5) & 92.4(+0.4)\\
\hline
\checkmark & \checkmark  & \checkmark & 92.3(+1.0) & 93.1(+1.1)\\
\hline
\end{tabular}
\end{center}
\caption{Comparison of top-1 accuracy(\%) on CUB-200-2011 with different module added to Swin Transformer.}
\label{tab2}
\end{table}

\begin{table}
\begin{center}
\begin{tabular}{|c|c|c|c|}
\hline
\multicolumn{3}{|c|}{Module} & \multicolumn{1}{|c|}{Backbone} \\
\hline
PA & Refinement & BS & ResNet-50\\
\hline
&  &  & 88.2 \\
\hline
\checkmark &  &  & 88.6(+0.4) \\
& \checkmark &  & 88.7(+0.5) \\
&  & \checkmark & 88.4(+0.2) \\
\hline
\checkmark & \checkmark & \checkmark & 89.8(+1.6) \\
\hline
\end{tabular}
\end{center}
\caption{Comparison of top-1 accuracy(\%) on CUB-200-2011 with different module added to ResNet-50.}
\label{tab3}
\end{table}

\subsection{Dataset and implement detail}
The datasets used in this study are the CUB200-2011\cite{CUB_200_2011} and NA-Birds\cite{NA-Birds}, two fine-grained bird classification datasets. The CUB200-2011 dataset has a total of 200 bird categories, including 5,994 training images and 5,794 testing data. Each category contains about 30 training and testing data. NA-Birds is larger than CUB200-2011, has 555 bird species, 23,929 training images and 24,633 test images. Both datasets provide image-level annotations and keypoint locations, but only image-level annotations will be used in this paper. When using ResNet-50\cite{ResNet} as the backbone network, the input image is a 448$\times$448 color image, and when using Swin-Transformer\cite{Swin-T}, the input image is a 384$\times$384 color image. The methods of data augmentation is as follows. If the input image size is 384×384, the first step is to scale the image to 510$\times$510, and if the input image size is 448$\times$448, it is scaled to 600$\times$600. In training phrase, data augmentation is performed via Randon Crop, Random HorizontalFlip, Random GaussianBlur, and Normalizarion while in testing phrase, Center Crop and Normalizarion is used. During training, the learning rate is set to 0.0005, with cosine decay and weight decay set to 0.0005. The optimizer used is SGD, with a batch size of 8, gradient accumulation steps set to 4, and the model is trained for a total of 80 epochs. All experiments are completed on a single Nvidia GeForce RTX 3090, and the Pytorch toolbox is used as the main implementation substrate. It takes about 5 hours to complete the training on CUB200-2011, and about 16 hours for NA-Birds.

\subsection{Ablation experiments}

\begin{figure*}
    \begin{center}
    \includegraphics[width=0.97\linewidth]{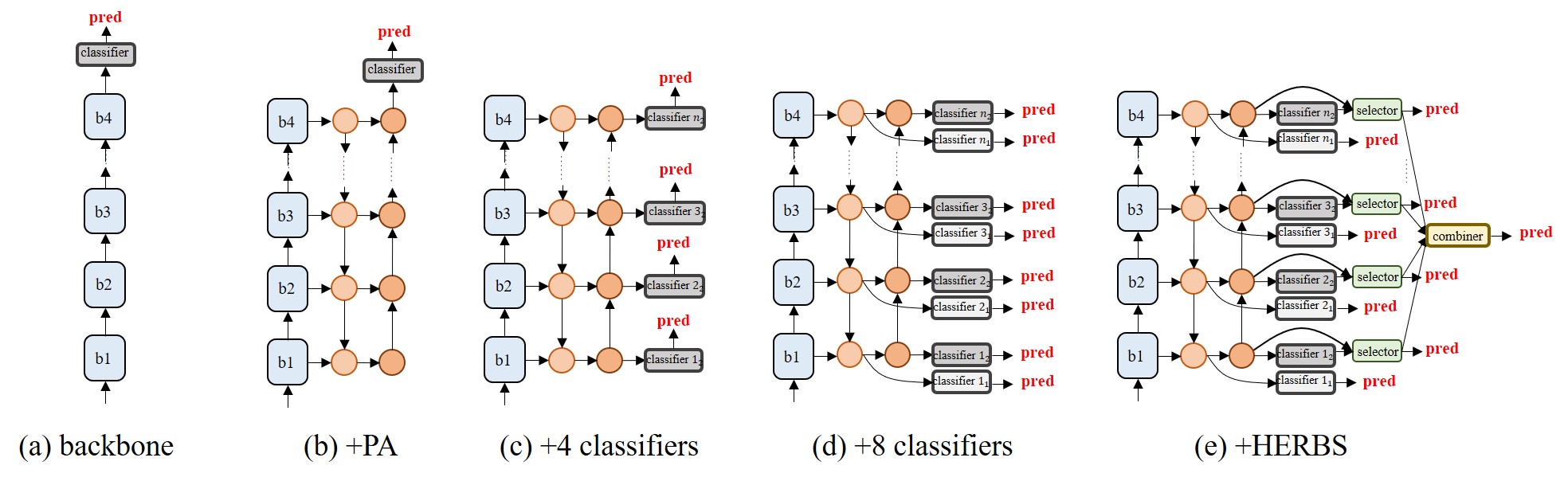}
    \end{center} 
\caption{The structure of models, (a) original backbone, the blue box represent the backbone blokcs. (b) backbone + path aggregation module. (c) backbone + PA module with four classifiers on the last bottom-up path. (d) backbone + PA module with eight classifiers on the top-down and bottom-up path. (e) backbone + HERBS}
\label{fig:5_modules}
\end{figure*}

\begin{figure*}
    \begin{center}
    \hspace{2cm}\includegraphics[width=0.98\linewidth]{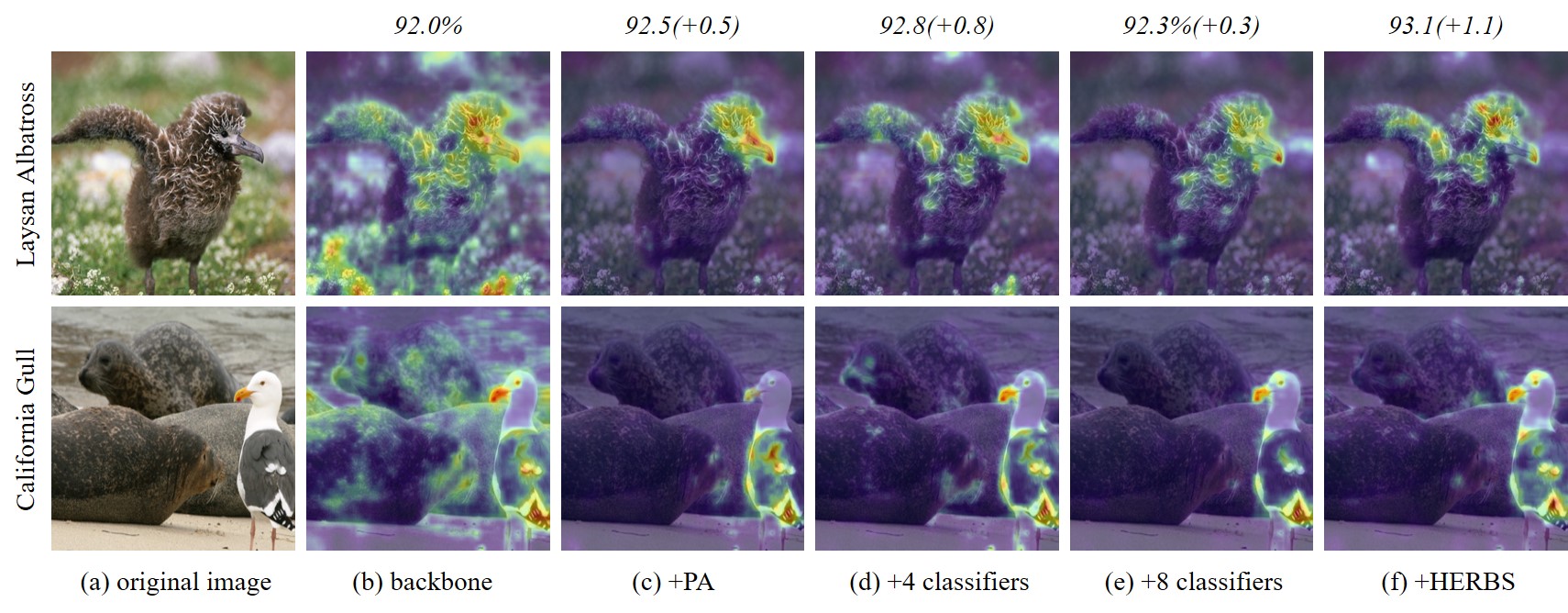}
    \end{center} 
\caption{Visualization of heat maps generated from different model. (a) original color image, (b) Swin Transformer backbone, (c) backbone + PA, (d) backbone + PA with four classifier, (e) backbone + PA with eight classifiers. (f) backbone + HERBS. The number on the top of the images represents the accuracy of the corresponding model.}
\label{fig:5_modules_heat}
\end{figure*}

In Table \ref{tab1}, we compare our proposed HERBS with state-of-the-art methods on CUB200-2011 and NA-Birds dataset. The middle column of Table \ref{tab1} shows that the proposed HERBS can reach 93.1\% in Top-1 accuracy, which is 0.7\% higher than the previous best method. Table \ref{tab1} last column shows that the proposed HERBS can reach 93.0\% in Top-1 accuracy on NA-Birds dataset, beating the previous state-of-the-art approaches. These results demonstrate that the proposed HERBS can effectively filter out background noise and extract appropriately sized discriminative features, enabling the identification of fine-grained categories accurately.

To better understand the impact of each module proposed in HERBS, we separately added the PA, Refinement, and BS modules to the classification backbone. First, the Swin Transformer Base (Swin-Base) and Swin Transformer Large (Swin-Large) were used as the testing backbone. As shown in Table\ref{tab2}, the original accuracies of Swin-Base and Swin-Large were 91.3\% and 92.0\%, respectively. After adding the PA, Refinement, or BS modules, there was a slight improvement in accuracy. The structure of only adding PA is shown in Fig.\ref{fig:5_modules}(b), only adding Refinement is demonstrated in Fig.\ref{fig:5_modules}(b), and only adding BS is shown in Fig.\ref{fig:5_modules}(a). The last row in Table 2 shows that the HERBS module improves the backbone's accuracy by about 1\%, demonstrating the module's effectiveness. 

The HERBS module can be utilized not only with transformer structures but also with convolution-based methods. We chose ResNet-50 as the test backbone, and the results of adding different modules are presented in Table \ref{tab3}. Interestingly, the HERBS module improves the accuracy of ResNet-50 ($+1.6$) more than the Swin Transformer ($+1.1$). This discrepancy may be attributed to the difference in input image resolutions, with ResNet-50 adapting to 448$\times$448 while the Swin Transformer only adapts to 384$\times$384. The resolution issue will be a topic for future discussion as it still needs to be addressed in this work. Generally speaking, the HERBS module demonstrates promising results on different types of backbones.

\textbf{About the capabilities of different model structures.} We further investigate on the ``growth'' and ``decline'' of receptive fields and present the five stages of our experiments. First, we tested the original backbone, as shown in Fig.\ref{fig:5_modules}(a), and the corresponding heat map is presented in Fig.\ref{fig:5_modules_heat}(b). It was observed that the model paid attention to a large amount of background area, indicating that the original backbone is not designed for detecting details in fine-grained data. 

Second, we added the features fusion module PA to the backbone, and the corresponding heat map is presented in Fig.\ref{fig:5_modules_heat}(c), with the structure depicted in Fig.\ref{fig:5_modules}(b). From the heat maps, we deduced that using only the last result, the response of the label could be focused on a small region. This improvement compared to the original model, however, still has a narrow focus. Next, we added classifiers to every block of the previous structure, as shown in Fig.\ref{fig:5_modules}(c). This structure widened the attention area, as shown in Fig.\ref{fig:5_modules_heat}(d), and effectively utilized multi-scale features. 

In the fourth step, we added four more classifiers to the model, as depicted in Fig.\ref{fig:5_modules}(d). These eight classifiers constrain the attention area, but the accuracy decreased. Finally, we added the HERBS module onto the backbone, and the corresponding heat map is presented in Fig.\ref{fig:5_modules_heat}(f). This module maintained detail while capturing a wide range of information, with the attention area approximately between Fig.\ref{fig:5_modules_heat}(d) and Fig.\ref{fig:5_modules_heat}(e). The results demonstrate that HERBS achieved better accuracy.

In this example, only HERBS predicted the image correctly, while the other models predicted the image to the wrong but visually similar class. This demonstrates that fine-grained visual classification requires detailed features rather than features that are too narrow.

\begin{table}
\begin{center}
\begin{tabular}{|c|c|c|c|c|c|}
\hline
\multicolumn{2}{|c|}{} & \multicolumn{2}{|c|}{Baseline} & \multicolumn{2}{|c|}{+HERBS}\\
\hline
Generic Class & Num. & Pr.($\uparrow$) & FP($\downarrow$)  & Pr.($\uparrow$) & FP($\downarrow$) \\
\hline
Flycatcher & 210 & 83.09 & 17 & 85.02 & 15\\
Gull & 240  & 80.79 & 1 & 81.66 & 1\\
Kingfisher & 150 & 93.33 & 1 & 94.67 & 0\\
Sparrow & 629 & 91.05 & 4 & 92.36 & 2\\
Tern & 209 & 75.12 & 0 & 83.73 & 0\\
Vireo & 210 & 92.46 & 10 & 94.47 & 8\\
Warbler & 750 & 94.73 & 13 & 95.54 & 14\\
Woodpecker & 179 & 97.63 & 0 & 97.63 & 0\\
Wren & 209 & 92.85 & 5 & 91.91 & 5\\
\hline
Average & 310 & 89.01 & 5.67 & \textbf{90.78} & \textbf{5.00}\\
\hline
\end{tabular}
\end{center}
\caption{Show the number (Num.) of generic classes, the precision (Pr.) (\%) and the number of false positives (FP) within the fine-classes in CUB-200-2011. The symbol $\uparrow$ denotes that a higher value is better, while $\downarrow$ denotes the opposite. We pick generic classes that contain more than six categories.}
\label{tab4}
\end{table}

\textbf{How the performance of HERBS on fine-classes?} We evaluated the performance of HERBS on real fine classes in the CUB-200-2011 dataset, which contains about 70 generic classes with 1 to 25 subcategories each, including 9 generic classes with more than 6 categories. Table \ref{tab4} lists these 9 generic classes, and we evaluate the models' performance on them.

The results show that HERBS outperforms the Swin Transformer baseline regarding both precision and false positive (FP) numbers. FP refers to cases where the model predicts a class that does not belong to the correct generic class. A lower FP number means that wrong predictions occur within the similar category, indicating that the model is not making serious mistakes. For example, doctors can trust the results from the fine-grained model and only need to check similar situations, reducing the effort required for double-checking.

\begin{figure}[t]
    \begin{center}
    \includegraphics[width=0.8\linewidth]{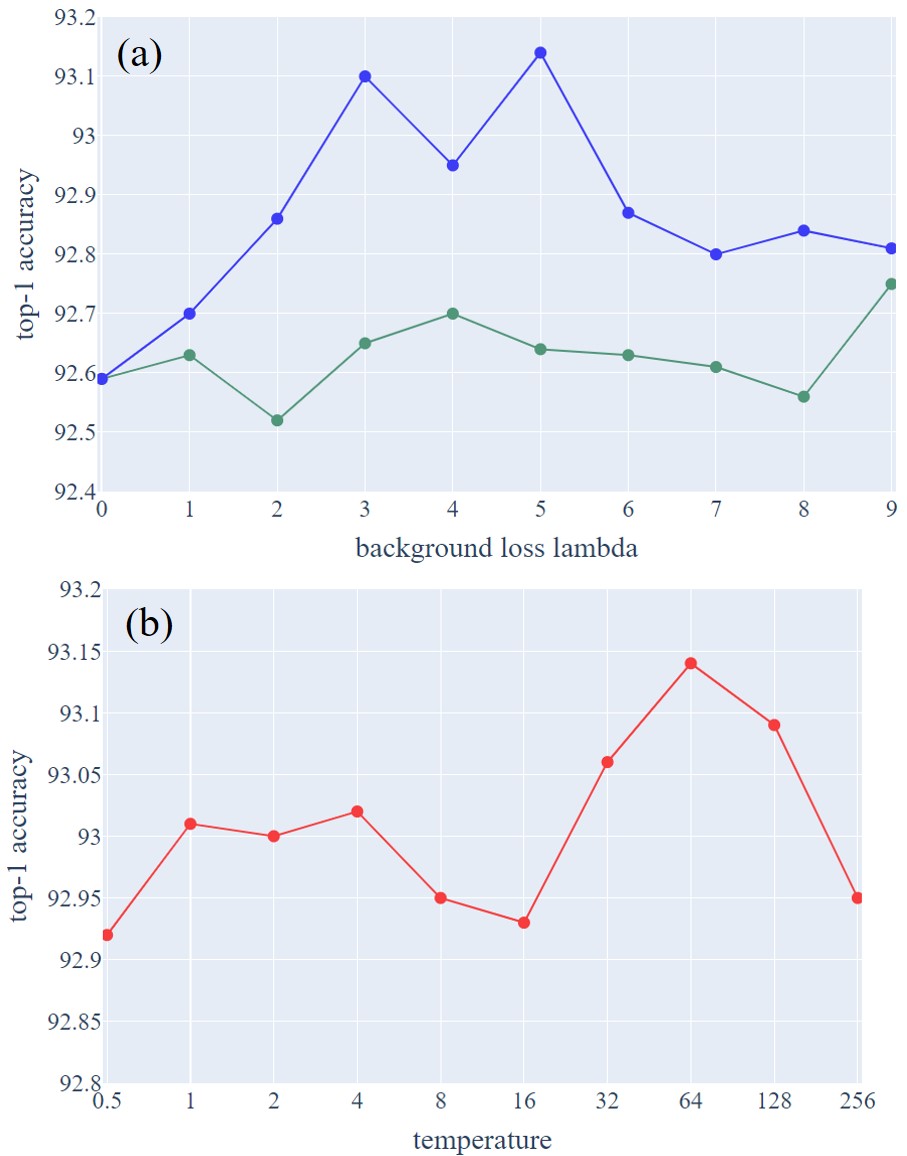}
    \end{center}
\caption{Comparison of top-1 accuracy with different hyperparameters. (a) shows the top-1 accuracy for different $\lambda_{d}$ values ranging from 0 to 9. The blue line represents the use of the tanh function, and the green line represents the use of the softmax function to map the classification results. (b) shows the top-1 accuracy for different temperatures ranging from 0.5 to 256.}
\label{fig:Temperature_and_BS_lambda}
\end{figure}

\begin{figure}
    \begin{center}
    \includegraphics[height=0.6\linewidth]{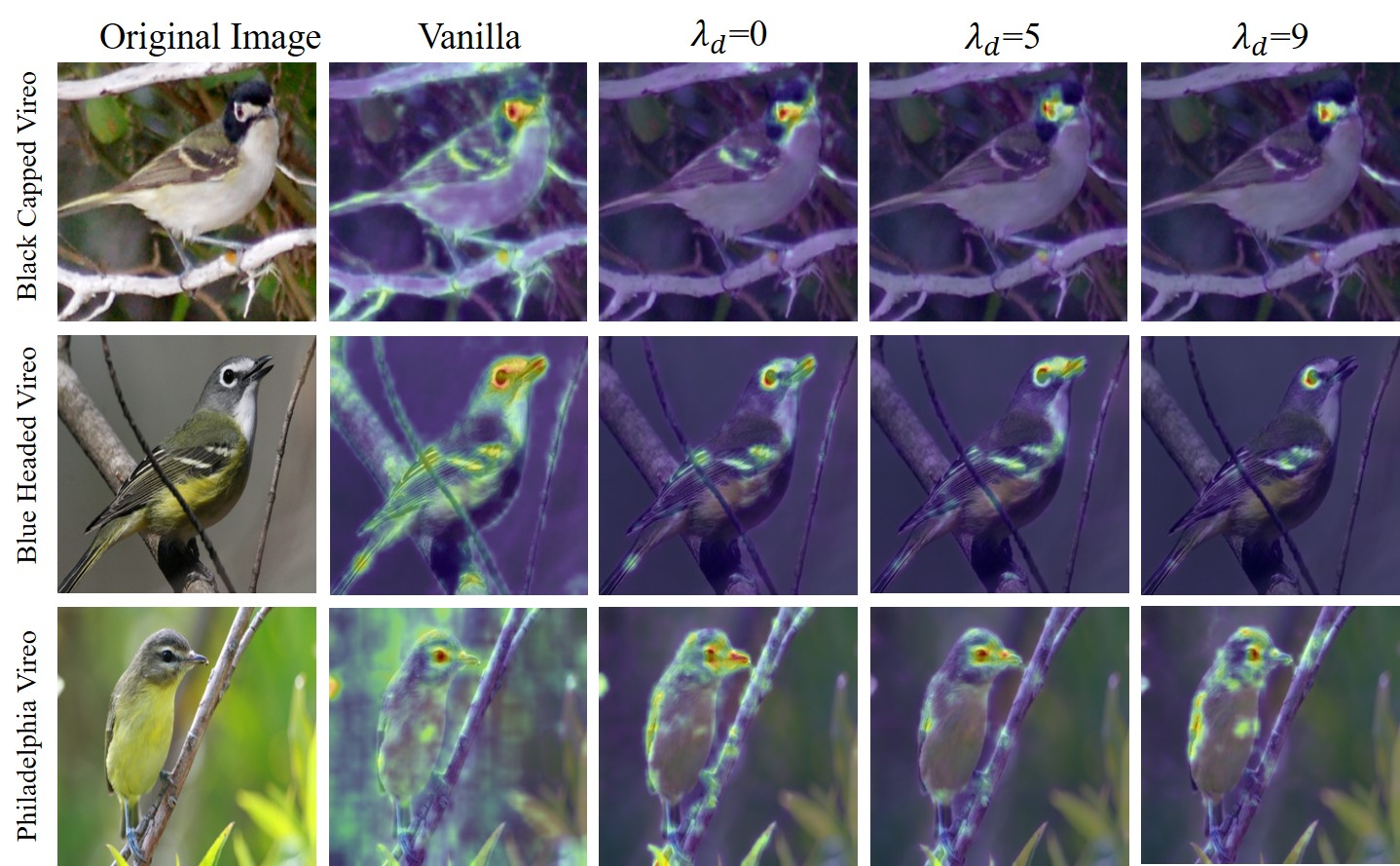}
    \end{center}
\caption{Visualization of the heat maps for different $\lambda_{d}$ values.}
\label{fig:BS_Heat}
\end{figure}

\begin{figure}[t]
    \begin{center}
    \includegraphics[height=0.95\linewidth]{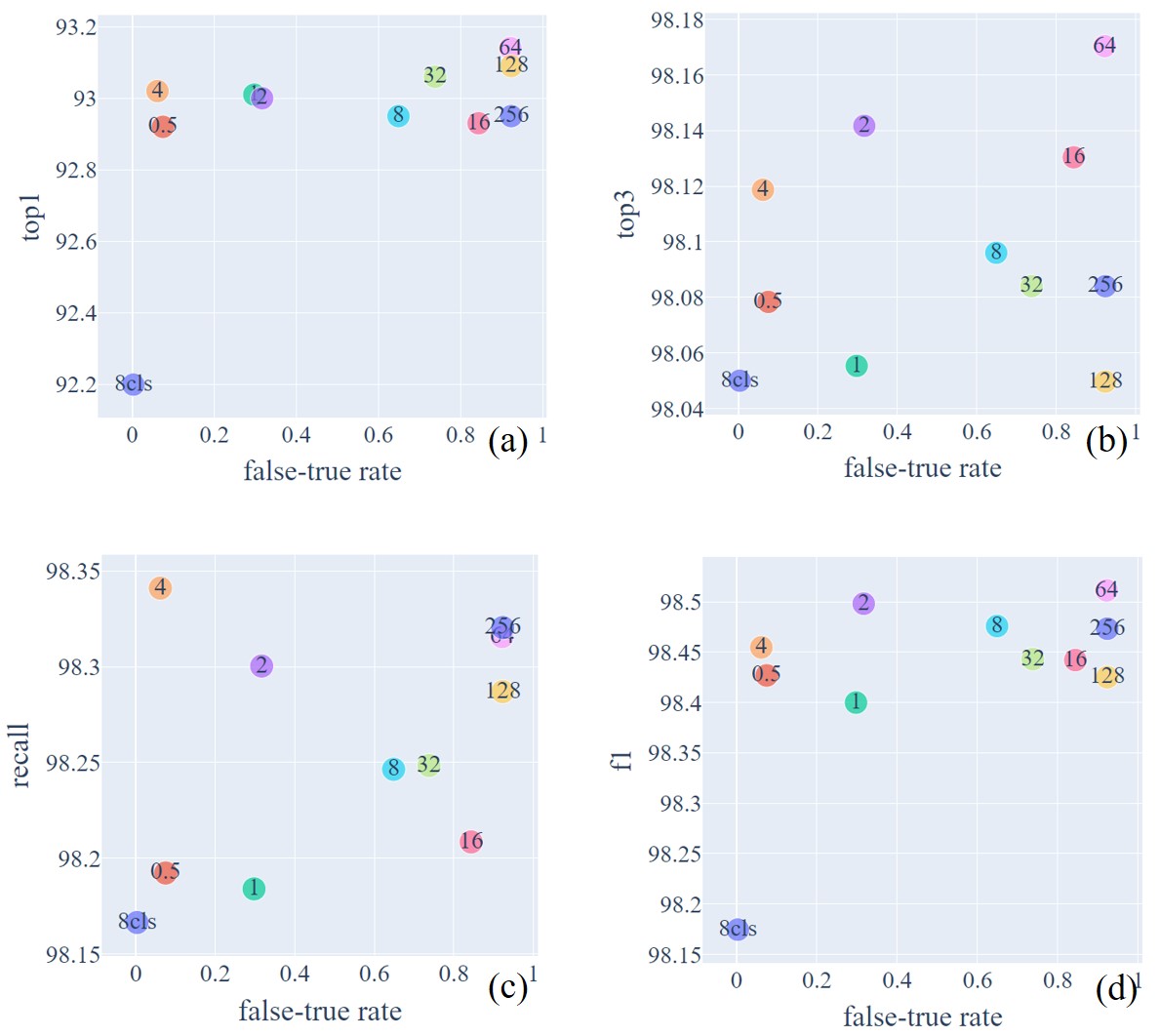}
    \end{center}
\caption{The False-True rate and its relation with (a) CUB-200-2011 top-1 accuracy, (b) top-3 accuracy, (c) the recall of fine-classes, (d) the f1 score of fine-classes.}
\label{fig:False-True}
\end{figure}

\textbf{How does the BS module suppress the background?} We tested the suppression intensity $\lambda_{d}$ from 0 to 9 and plotted their corresponding top-1 accuracy in Fig.\ref{fig:Temperature_and_BS_lambda}(a) (blue line). The corresponding heat maps for different $\lambda_{d}$ values are shown in Fig.\ref{fig:BS_Heat}. From the heat maps, we observed that when $\lambda_{d}$ is set to 0, which means only the merged loss (selected areas) is used to constrain the feature map, the model still pays attention to some background areas. Comparing the heat maps of $\lambda_{d}$ at 0, 5, and 9, we observed that the concentration level increases as the suppression intensity $\lambda_{d}$ increases, demonstrating that the BS module can effectively suppress background values. 

As mentioned before, the tanh function is used to map the classification results instead of the softmax function. In Fig.\ref{fig:Temperature_and_BS_lambda}(a), the blue line represents the tanh-based approach, while the green line represents the softmax-based approach. In the softmax-based method, the pseudo-target would be $\frac{1}{C_{gt}}$. However, this may lead to unstable training because even though we refer to the noisy or unselected area as "background", it is not necessarily the same as background elements such as the sky, trees, or ocean. Some unselected areas may still appear on the bird's body and could be present in other categories' appearances. Therefore, it is crucial to separate them by feature values rather than class probability. The impact of suppression intensity is shown in Fig. \ref{fig:BS_Heat}.

\textbf{What does high-temperature provide?} We emphasize that the use of high-temperature is based on our experiment results, as shown in Fig.\ref{fig:Temperature_and_BS_lambda}(b), where we discovered that the best top-1 accuracy and top-3 accuracy occurred at a temperature of 64. A high temperature would cause the distribution to become very flat, which means that the model has higher tolerance for misclassification. This tolerance allows the model to discover more diverse features and use multi-class features to enhance its capability. 

To explain this further, we present the impactness of false-true rates in Fig.\ref{fig:False-True}. Here, false-true represents the number of wrong predictions in the top-bottom path but correct in the bottom-up path. The false-true rate is calculated using the following equation:
\begin{equation} \label{eq:herbs_loss}
\text{false-true rate} = \frac{\text{false-true}}{(\text{false-true} + \text{false-false)}}
\end{equation}

The higher false-rate means the model is not only focus on one target. We discovered that the top-1 and top-3 accuracy of the dataset, as well as the F1 score of the fine-classes, are slightly related to this. The structure used in the 8cls dot in Fig.\ref{fig:False-True} is Fig.\ref{fig:5_modules}(d), while the others are HERBS with different temperatures.

\section{Conclusion}
In this paper, we proposed HERBS with the BS module and the high-temperature refinement module which can be easily applied to popular backbone networks. The method effectively filters out background noise and focuses on discriminative features while maintaining a proper attention area scale. Our experiments on fine-grained visual classification tasks show that HERBS significantly improves accuracy and outperforms state-of-the-art methods on the CUB-200-2011 and NA-Birds benchmark datasets. Future work can explore the use of adaptive strategies to choose the temperature or suppression intensity and investigate low computation cost methods based on this work. Overall, the proposed HERBS can achieve high accuracy up to 93\% to provide a promising solution to improve the performance of fine-grained visual classification tasks.

{\small
\bibliographystyle{ieee_fullname}
\bibliography{egbib}
}

\end{document}